\documentclass[runningheads]{llncs}
\pdfoutput=1
\usepackage{graphicx}
\usepackage{amsmath,amssymb,amsfonts, easyReview}
\usepackage{caption}
\usepackage{subcaption}
\usepackage{bbding}
\usepackage{adjustbox}
\usepackage{graphbox}
\usepackage[marginal]{footmisc}

\begin{document}
\captionsetup[figure]{labelfont={bf},labelformat={default},labelsep=period,name={Fig.}}
\captionsetup[table]{labelfont={bf},labelformat={default},labelsep=period,name={Table.}}
\title{Mixline: A Hybrid Reinforcement Learning Framework for Long-horizon Bimanual \\Coffee Stirring Task} 
%
%
\author{Zheng Sun\inst{^1} \and
Zhiqi Wang\inst{^1} \and
Junjia Liu\inst{^1} \and
Miao Li\inst{^2}\inst{(}\Envelope\inst{)} \and
Fei Chen\inst{^1}\inst{(}\Envelope\inst{)}}
\authorrunning{Zheng, Zhiqi, et al.}
\titlerunning{Long-horizon Bimanual Coffee Stirring}
%
\institute{
The Chinese University of Hong Kong, Hong Kong, China \email{f.chen@ieee.org} \and
Wuhan University, Wuhan, China \email{miao.li@whu.edu.cn}
} 

\maketitle              
\footnote{$^*$ First Author and Second Author contribute equally to this work. \\
$^*$ This work was supported in part by the Research Grants Council of the Hong Kong Special Administrative Region, China under Grant 24209021, in part by the VC Fund of the CUHK T Stone Robotics Institute under Grant 4930745, in part by CUHK Direct Grant for Research under Grant 4055140, and in part by the Hong Kong Centre for Logistics Robotics.\\}

\begin{abstract}

Bimanual activities like coffee stirring, which require coordination of dual arms, are common in daily life and intractable to learn by robots. Adopting reinforcement learning to learn these tasks is a promising topic since it enables the robot to explore how dual arms coordinate together to accomplish the same task. However, this field has two main challenges: coordination mechanism and long-horizon task decomposition. Therefore, we propose the \textit{Mixline} method to learn sub-tasks separately via the online algorithm and then compose them together based on the generated data through the offline algorithm. We constructed a learning environment based on the GPU-accelerated Isaac Gym. In our work, the bimanual robot successfully learned to grasp, hold and lift the spoon and cup, insert them together and stir the coffee. The proposed method has the potential to be extended to other long-horizon bimanual tasks.


\keywords{Reinforcement learning  \and Bimanual coordination \and Isaac Gym}
\end{abstract}
\section{Introduction}
	 	 \par
\begin{figure}
		\centering
		\includegraphics[width=0.5\linewidth]{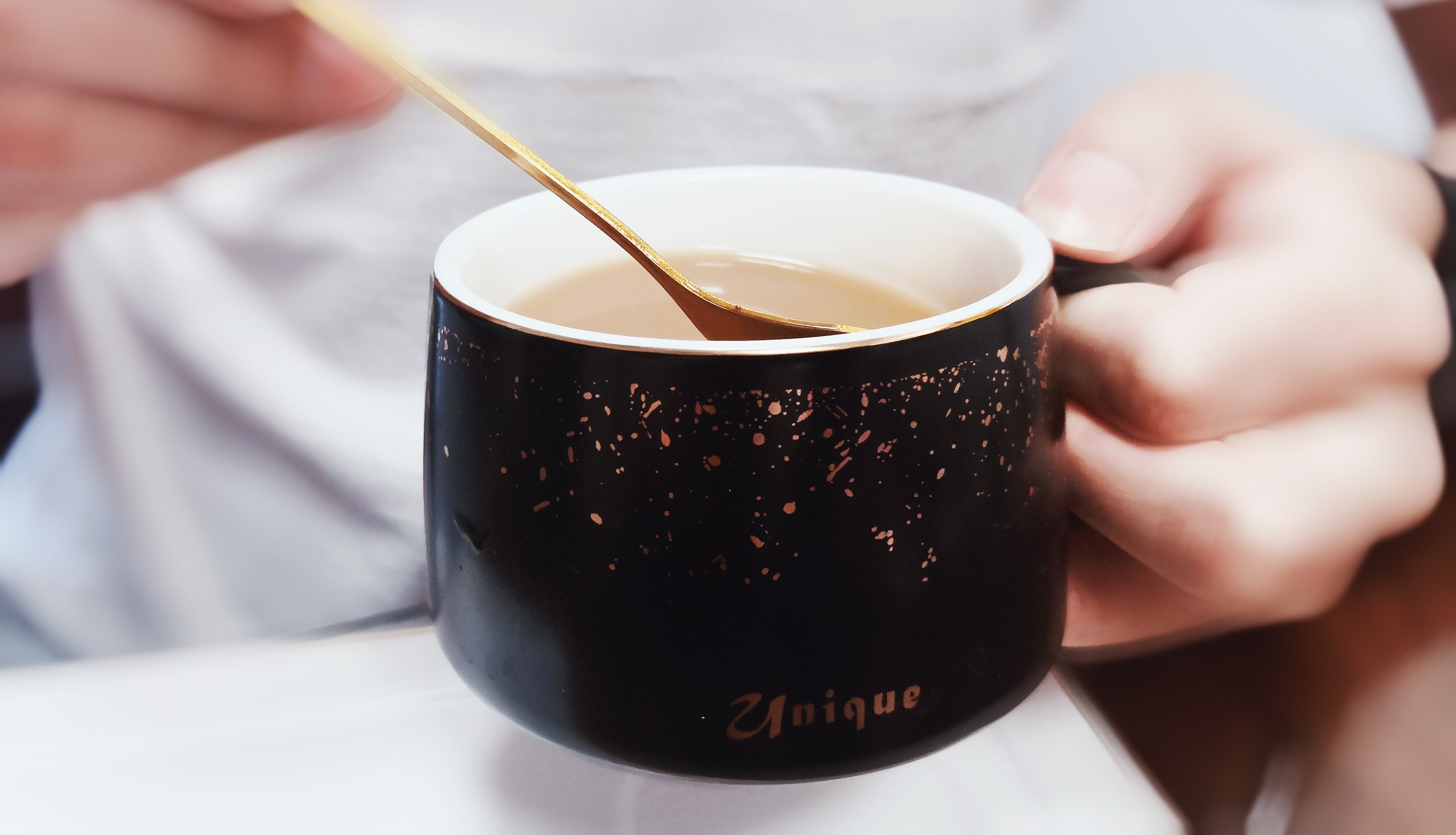}
		\caption{Bimanual coffee stirring}  
		\label{Bimanual coffee stirring}
	\end{figure}

 The rapid development of Reinforcement Learning (RL) has provided new ideas for robot control \cite{num1}\cite{num2}, and the training of bimanual robots with coordination has become a hot topic in reinforcement learning. In this paper, we are interested in how to make bimanual robots learn human's daily activities. These activities are usually long-horizon and need the coordination of dual human arms. These two challenges limit the use of RL in complex tasks and make it unavailable to real-world scenarios. Here we design a long-horizon and bimanual coffee stirring task as an example to study the potential solution of these kinds of tasks. The illustration of coffee stirring movements is shown in Fig. \ref{Bimanual coffee stirring}. It contains movements like grasping, lifting, inserting, and stirring. 
	 	
	 	 
    

	 In recent research, Zhang et al. proposed a novel disentangled attention technique, which provides intrinsic regularization for two robots to focus on separate sub-tasks and objects \cite{num3}. Chitnis et al. decomposed the learning process into a state-independent task schema to yield significant improvements in sample efficiency \cite{num4}. Liu et al. applied dual-arm Deep Deterministic Policy Gradient (DDPG) \cite{num5} based on Deep Reinforcement Learning (DRL) with Hindsight Experience Replay (HER) \cite{num6} to achieve cooperative tasks based on “rewarding cooperation and punishing competition” \cite{num7}. Chiu et al. achieved bimanual regrasping by using demonstrations from a sampling-based motion planning algorithm and generalizing for non-specific trajectory \cite{num8}. Rajeswaran et al. augmented the policy search process with a small number of human demonstrations \cite{num9}. Liu et al. proposed a novel scalable multi-agent reinforcement methods for solving traffic signal control in large road networks \cite{NUM1}. Cabi et al. introduced reward sketching to make a massive annotated dataset for batch reinforcement learning \cite{num10}. Mandlekar et al. presented Generalization Through Imitation (GTI), using a two-stage intersecting structure to train policies \cite{num11}. Zhang et al. gave a deep analysis of Multi-Agent RL theory and assessment of research directions \cite{num12}. Except for adopting multi-agent RL methods, some recent work also focus on learning bimanual complex robot skills from human demonstration \cite{num13}\cite{NUM3}. Specially, Liu et al. achieved bimanual robot cooking with stir-fry, the traditional Chinese cooking art. They combined Transformer with Graph neural network as \textit{Structured-Transformer} to learn the spatio-temporal relationship of dual arms from human demonstration \cite{num13}.

      In this paper ,we propose to use a task division process first to decompose the long-horizon task and learn each sub-task separately. Besides, we propose wait training method to solve the poor coordination. Then by adopting the conservative Q-learning (CQL), we can combine the offline data generated via the separate learning process to achieve the learning of the whole task. We regard this hybrid reinforcement learning method which contains both online and offline RL algorithms, as \textit{Mixline}. 

	\section{Task definition and environment}

	\subsection{Task definition}
	When we drink coffee, we pick up the spoon and the coffee cup with our left and right hands and use the spoon to stir the coffee in the cup. Therefore, the goal is to train two robotic arms to do the same sequential actions: one arm grasps a spoon, the other grasps a cup, then the arm with the spoon lifts and inserts it into the cup for a stirring motion.

	We found that the previous work still has flaws, such as bimanual coordination and long-horizon tasks. These are common challenges in the robot field. To solve them, in our study, we split the whole task into three stages. By setting the reward function for the three stages separately, we solve the existing sparse reward problem. Splitting into three stages helps us solve the long-horizon and poor coordination problems.  
	

    The three stages are:
    \begin{enumerate}
        \item Grasp and pick up the cup and spoon;
		\item Insert the spoon into the cup;
		\item Stir in the cup without hitting the cup overly.
    \end{enumerate}
    
\begin{figure*}[htp]
        \centering
         \begin{subfigure}[b]{0.3\textwidth}
             \centering
             \includegraphics[width=\textwidth]{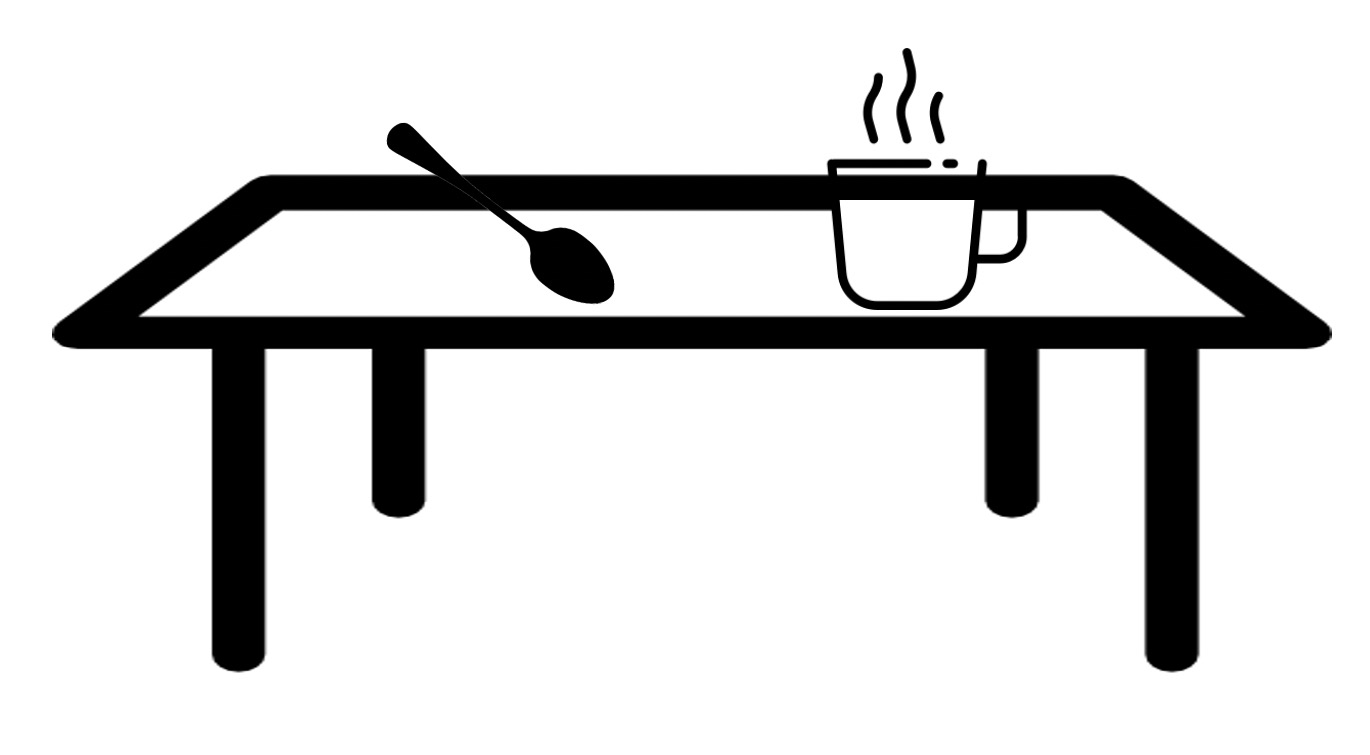}
            \caption{Hold and lift}
            \label{taska}
         \end{subfigure}
         \hfill
         \begin{subfigure}[b]{0.3\textwidth}
             \centering
             \includegraphics[width=\textwidth]{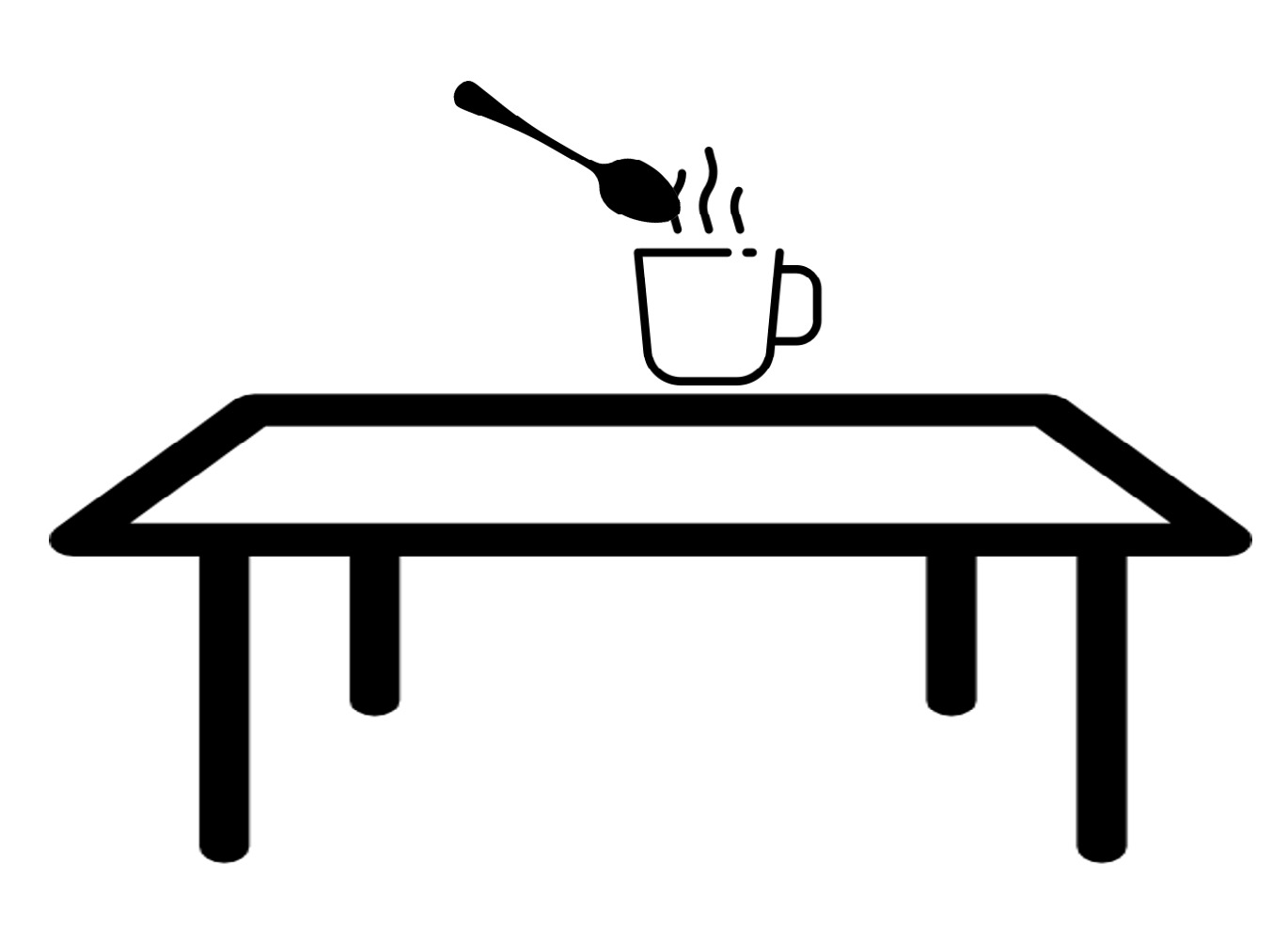}
            \caption{Insert}
        \label{taskb}
         \end{subfigure}
         \hfill
         \begin{subfigure}[b]{0.3\textwidth}
             \centering
             \includegraphics[width=\textwidth]{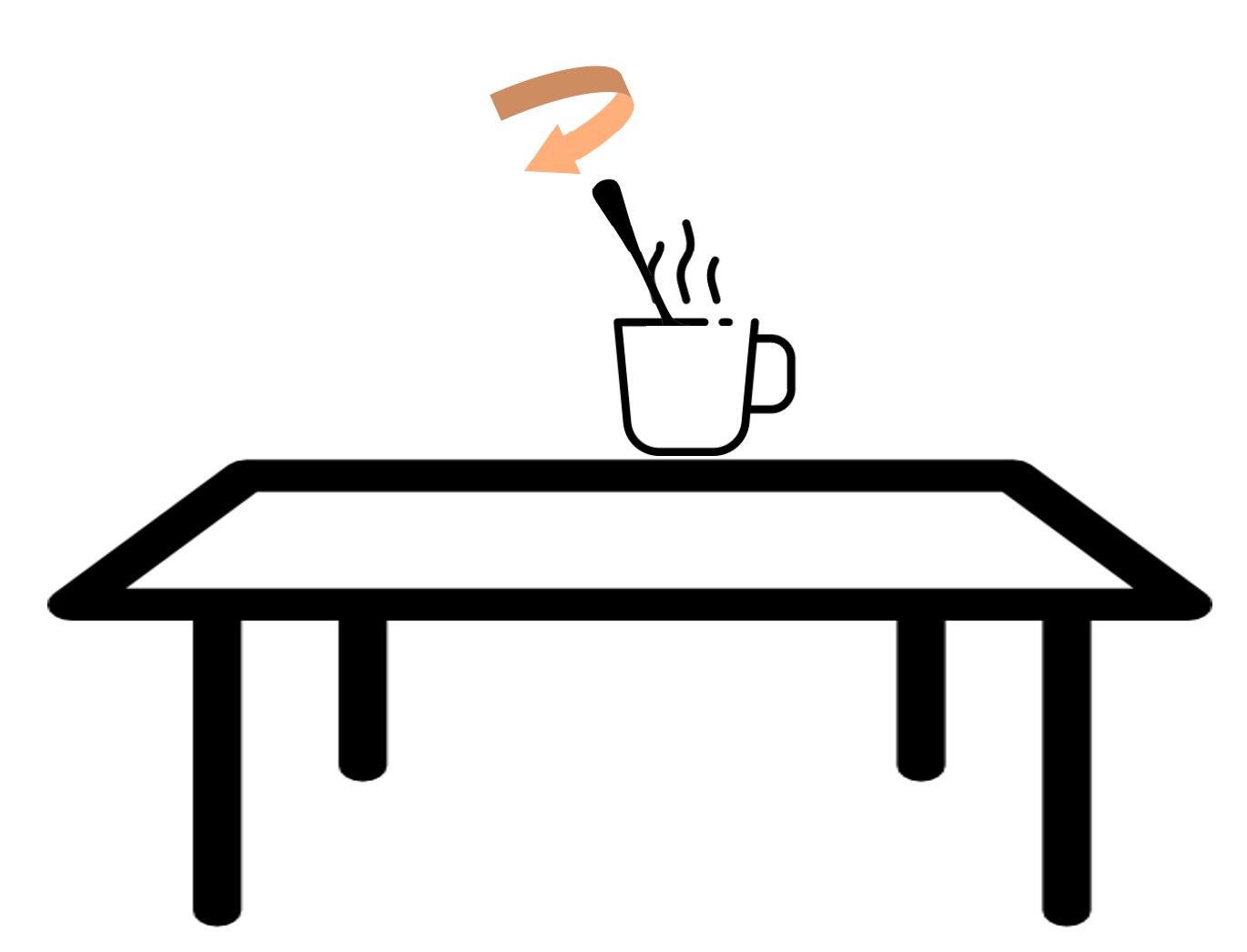}
            \caption{Stir overly}
        \label{taskc}
         \end{subfigure} \caption{Separate the coffee stirring task into three stages. From (a) to (c), there are diagrams about stage 1, stage 2, and stage 3.}\label{coffee_stir}
        \end{figure*}

	The diagrams about three stages are shown in Fig. \ref{coffee_stir}. 
	
    
	\subsection{Isaac gym environment}
	Isaac Gym offers a high-performance learning platform to train policies for a wide variety of robotic tasks directly on GPU \cite{num14}. It performs significantly better than other standard physics simulators using CPU clusters, and also has many advantages in reinforcement learning:
    \begin{itemize}
		\item It provides a high-fidelity GPU-accelerated robot simulator platform;
		\item With the help of Tensor API, Isaac Gym goes through CPU bottlenecks by wrapping physics buffers into PyTorch \cite{num15} tensors;
		\item It can achieve tens of thousands of simultaneous environments on a single GPU \cite{num16}.
	\end{itemize}

	\section{Methodology}
	\subsection{Background of reinforcement learning}
	    
	    \par The dynamical system of reinforcement learning is defined by a Markov Decision Process (MDP) \cite{num17}. MDP can be regarded as a 5-tuple $M=(S,A,T,r,\gamma)$, where $S$ is states of the environment and agent, $A$ refers to the actions of the agent, $T$ defines the state transition function of the system, $r$ is the reward function, and $\gamma$ means the discount factor.
	    \par The goal of reinforcement learning is to learn a policy $\pi(a,s)$. And the objective $J(\pi)$ can be derived as an expectation under the trajectory distribution, as Equ. \ref{rlobjective}. 
	    \begin{equation} 
            J(\pi)=\mathbb{E}_{\tau\sim p_{\pi}(\tau)}[\sum_{t=0}^{H}\gamma^t r(s_t,a_t)]
         \label{rlobjective}
    	\end{equation}
    	where $\tau$ is the whole trajectory given by sequence of states $s$ and actions $a$, and $p_{\pi}(\tau)$ is the trajectory distribution of the given policy $\pi$.

    \subsection{Proximal policy optimization}
	    Proximal Policy Optimization (PPO) \cite{num18} is an on-policy algorithm, which is the most widely used algorithm in RL. Its the objective function is usually modified by important sampling \cite{num19}. Suppose the target policy is $\theta$ and the behavior policy is $\theta’$. Important sampling adopts KL divergence to measure the difference between these two policies and minimizes their gap. The modified objective function is
    

    	 \begin{equation} 
			\hat{J'}_{PPO}(\theta)=\hat{E}_{\pi}[{\frac{\pi_{\theta}(a_t|s_t)}{\pi_{\theta'}(a_t|s_t)}}{A^{\theta'}}{(s,a)}]-\beta{KL(\theta{,}\theta^{'})}
		 \label{cond1}
    	\end{equation}
\subsection{Wait training mechanism for bimanual learning}
We introduce a new training method to help us solve the problem of bimanual coordination, called wait training. The specific principle is that only one robotic arm is considered each time we train. Taking the first stage as an example, we lock the movement of the right arm so that the algorithm only focuses on the grasping of the spoon by the left arm. When the left arm grabs the spoon, lock the movement of the left arm. After training each arm to grasp successfully, unlock the arms movement and focus the training process on lifting the arms. Through wait training, we have solved the problem of poor coordination between arms under the PPO algorithm.
	\subsection{Mixline method for learning long-horizon tasks}
	    
	    After we finish the policy learning of each stage using PPO, we derive the bimanual trajectory and combine them to form the whole long-horizon trajectory. The combined trajectory is regarded as an expert dataset and used for offline training.
        \par To utilize offline data, we introduce conservative Q-learning (CQL) \cite{num20}, which can be implemented on top of soft actor-critic (SAC) \cite{num21}. Similar work has been proposed based on SAC to reduce the state-action space and improve the training efficiency by introducing inexplicit prior knowledge \cite{NUM2}. The primary purpose of CQL is to augment the standard Bellman error objective with a simple Q-value regularizer. In the experiment, it is proved effective in mitigating distributional shift compared to other offline algorithms.
        \par The conservative policy evaluation minimizes the Q value of all actions, then adds a term to maximize the Q value of actions from the expert dataset to encourage actions that conform to the offline dataset and restrain actions beyond the dataset, where Equ. \ref{cond1}. 
        \begin{equation} 
            \begin{aligned}
            \hat{Q}^{k+1}\gets \mathop{\arg\min_Q}\,\alpha\cdot(\mathbb{E}_{s\sim D,a\sim \mu(a|s)}[Q(s,a)]
            -\mathbb{E}_{s\sim D,a\sim \hat{\pi}_\beta(a|s)}[Q(s,a)])
            \\+\frac{1}{2}\mathbb{E}_{s,a,s^{'}\sim D}[(Q(s,a)-\hat{\mathcal{B}}^\pi \hat{Q}^k(s,a))^2]
            \end{aligned}
         \label{cond1}
    	\end{equation}
    	
    	By combining online PPO algorithm and offline CQL algorithm, the proposed \textit{Mixline} method can gather offline data generated from each stage and achieve the learning of the whole long-horizon tasks.

	\section{Experiments}
	\subsection{Simulation environment}
	    In our work, we set seven objects in each working space in the Isaac Gym environment. There are two Franka robots, one spoon on a shelf, one cup on a box shelf, and a table, as shown in the Fig. \ref{gymenv}. 
	    \subsubsection{Simulation:} 
\begin{figure*}[htp]
        \centering
         \begin{subfigure}[b]{0.4\textwidth}
         \centering
    		\includegraphics[width=0.8\linewidth]{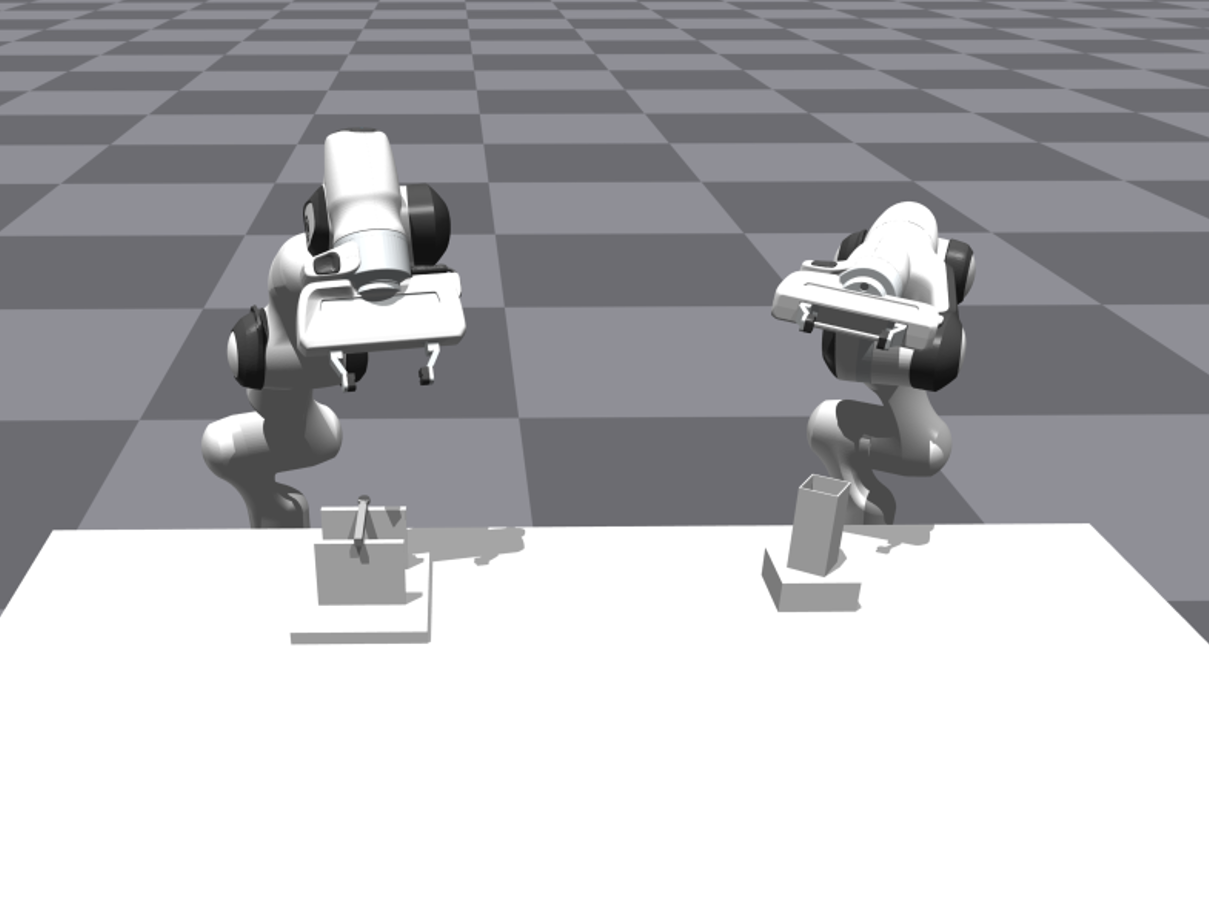}
    		\caption{Franka robots and assets in Isaac Gym environment} 
    		\label{gymenv}
    	\end{subfigure}
        \begin{subfigure}[b]{0.5\textwidth}
             \centering
    		\includegraphics[width=0.8\linewidth]{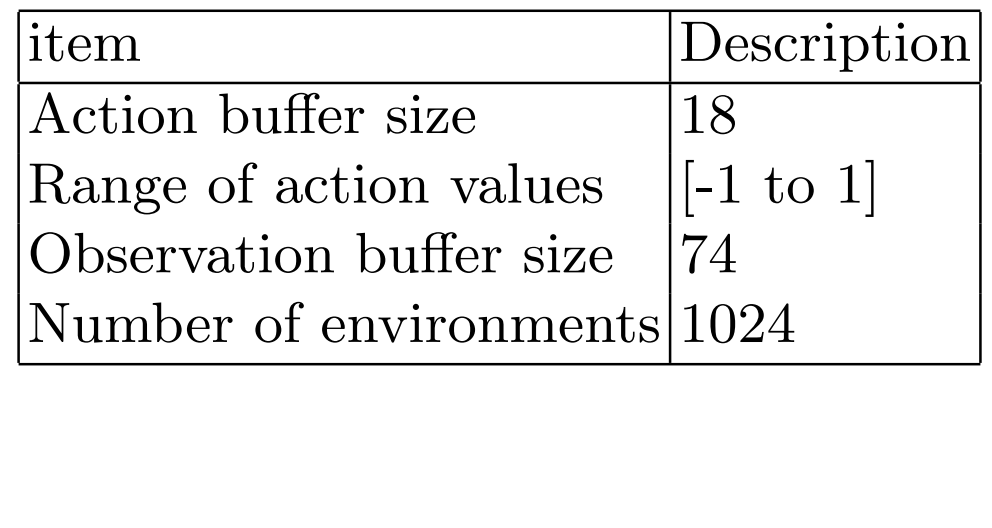}
    	 \caption{RL settings}\label{rlsettings}
        \end{subfigure}
     \caption{Simulation environment}\label{Simulation environment}
\end{figure*}

Each Franka robot has seven revolute joints and two prismatic joints in the gripper. The action space is set continuously between -1 to 1 and uses relative control. Some basic settings are shown in Fig. \ref{rlsettings}. To facilitate the state of the grasp in the training process, we divided the observation space in each workspace into 74. The Table \ref{Obversation buffer table} records the allocation of each part of the buffer. We labeled one Franka arm as Franka-spoon and the other as Franka-cup.
	    
	    
 
\subsubsection{RL flow:} In our experiment,we use PyTorch \cite{num15}\cite{num22} as base RL library. In the RL iteration, the agents sample actions from the policy and give the actions to the environment for physics simulation; then, the environment gives back all the buffer needed to compute for further steps such as reward calculation.

	 \begin{table}[ht]
         \centering
         \caption{Observation buffer table}\label{Obversation buffer table}
         \begin{tabular}{|l|l|}
            \hline
            index& Description\\
            \hline
            0-8& the scale of Franka-spoon’s joint positions\\
            9-17&the scale of Franka-cup’s joint positions \\
            18-26&the velocity of Franka-spoon’s joints\\
            27-29&the relative position between the top of gripper and spoon\\
            30-38&the velocity of Franka-cup’s joints\\
            39-41&the relative position between the top of gripper and cup\\
            42-48&spoon’s position and rotation\\
            49-55&cup’s position and rotation\\
            56-64&Franka-spoon’s joint positions\\
            65-73&Franka-cup’s joint positions\\
            \hline
        \end{tabular}
    \end{table}
    
	\subsection{Design of reward function}
	    In general, all the rewards and penalties are given a scale to get the total reward.
	    \\ \hspace*{\fill} \\
	\noindent\textbf{Reward}
	
\noindent We divide total rewards into Distance reward, Rotation reward, Around reward, Finger distance reward, and Lift reward for finishing the task.
	\paragraph{Distance reward:}
    	Distance reward is relative to the grasping point. The reward lifts when the robot hand moves towards the object's grasp point.
   \paragraph{Rotation reward:} 
    	We pre-define the axis to ensure alignment. The rotation reward gets its maximum value when the specified robot plane coincides with the pre-defined plane.
   
   \paragraph{Around reward:}
        When the robot grasps something, we have to make two fingers at the different sides of the object, not only orientation and distance. Only in this situation can the object be taken up. This reward is the prerequisite for taking up the object.

 \paragraph{Finger distance reward:}
       If the around reward is satisfied, it can grasp the object when the gripper is close. Meantime, the distance should be small to ensure both the finger and robot hand is close enough, and the gripper holds tighter, the reward is bigger. 

  \paragraph{Lift reward:}
        The target of stage 1 is to take up the object. Hence the reward boosts when the robot hand grasps the object successfully and lifts the object. In the meantime, we expect the gripper holds the object tightly so that the object will not drop during fast movement.\\ \hspace*{\fill} \\
    

\noindent\textbf{Penalty}\\

	    \noindent In general, the penalty term prevents the agent do unexpected actions. We divide total penalty into Action penalty, Collision penalty, Fall penalty, and Wrong pose penalty.
	    
  \paragraph{Action penalty:}
        Making a regularization of the actions means more actions and more penalties. We expect the robot can finish the task as soon as possible.
    \paragraph{Collision penalty:}
        The robot may hit the table if the link contact force exceeds the limit. An unexpected collision gives a penalty to the total reward.
   \paragraph{Fall penalty:}
        The robot receives a fall penalty if the object is knocked down by the robot arm when it tries to grasp or in other similar circumstances. 
   \paragraph{Wrong pose penalty:}
        The goal is to let the robots achieve bimanual coordination so that the trajectory should always be reasonable, not only the final target position. For example, the cup should not be inverted during the movement. Fall penalty can be added to this penalty.
	\subsection{Reset condition}
	
	The robot should always stay in its safe operational space. We do not expect the robot to be close to its singularity position. We reset the environment if the pose is unreasonable. For example, in our task, we hope the robot should not let the object drop; if it drops on the table or ground, we directly reset it and penalize it. In common, the environment should be reset if the task step exceeds the maximum task length.



	\subsection{Results}
\begin{figure*}[htp]
        \centering
         \begin{subfigure}[b]{0.3\textwidth}
             \centering
             \includegraphics[width=\textwidth]{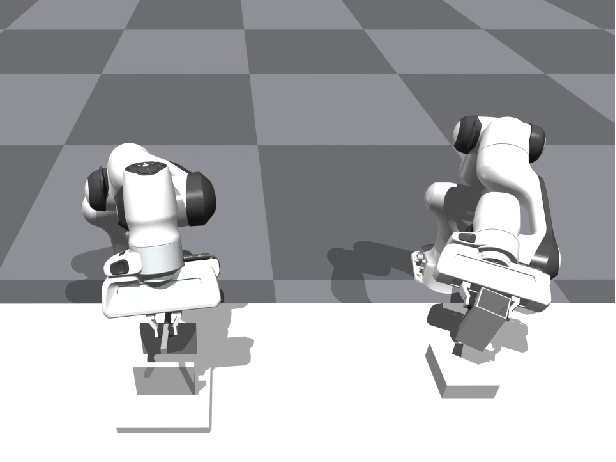}
             \caption{Stage 1: Hold and lift}  
		    \label{stage1}
         \end{subfigure}
         \hfill
         \begin{subfigure}[b]{0.3\textwidth}
             \centering
             \includegraphics[width=\textwidth]{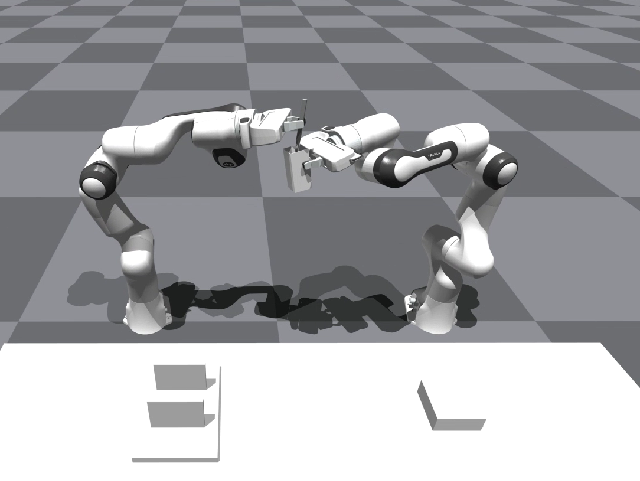}
             \caption{Stage 2: Insert}  
		\label{stage2}
         \end{subfigure}
         \hfill
         \begin{subfigure}[b]{0.3\textwidth}
             \centering
             \includegraphics[width=\textwidth]{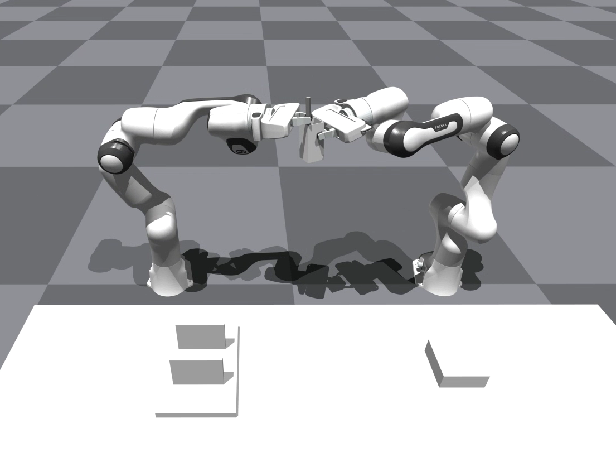}
             \caption{Stage 3: Stir overly}  
		\label{stage3}
         \end{subfigure} \caption{Result of each stage} \label{stageresult}
    \end{figure*}

    	In our work, we generated expert datasets with an online algorithm, then proposed combining the sub-task offline dataset with training the policy.
	    In stage 1, we trained to grasp cup and spoon in the PPO algorithm, used the wait training method to solve coordination problems, and the CQL+PPO method in several different environment settings. In stages 2 \& 3, we set the end of stage 1 as the initial state and trained the robot arm to complete the task. With the help of the above methods, We finish our training goals and results as shown in Fig.\ref{stageresult}. After combining all stages, we proposed to train the policy using the offline algorithm to finish the whole long-horizon task. The training results are consistent with our expectations, proving that the method proposed in bimanual reinforcement learning is effective.
	   
	\subsection{Ablation study}
\begin{figure*}[htp]
        \centering
         \begin{subfigure}[b]{0.3\textwidth}
             \centering
             \includegraphics[width=\textwidth]{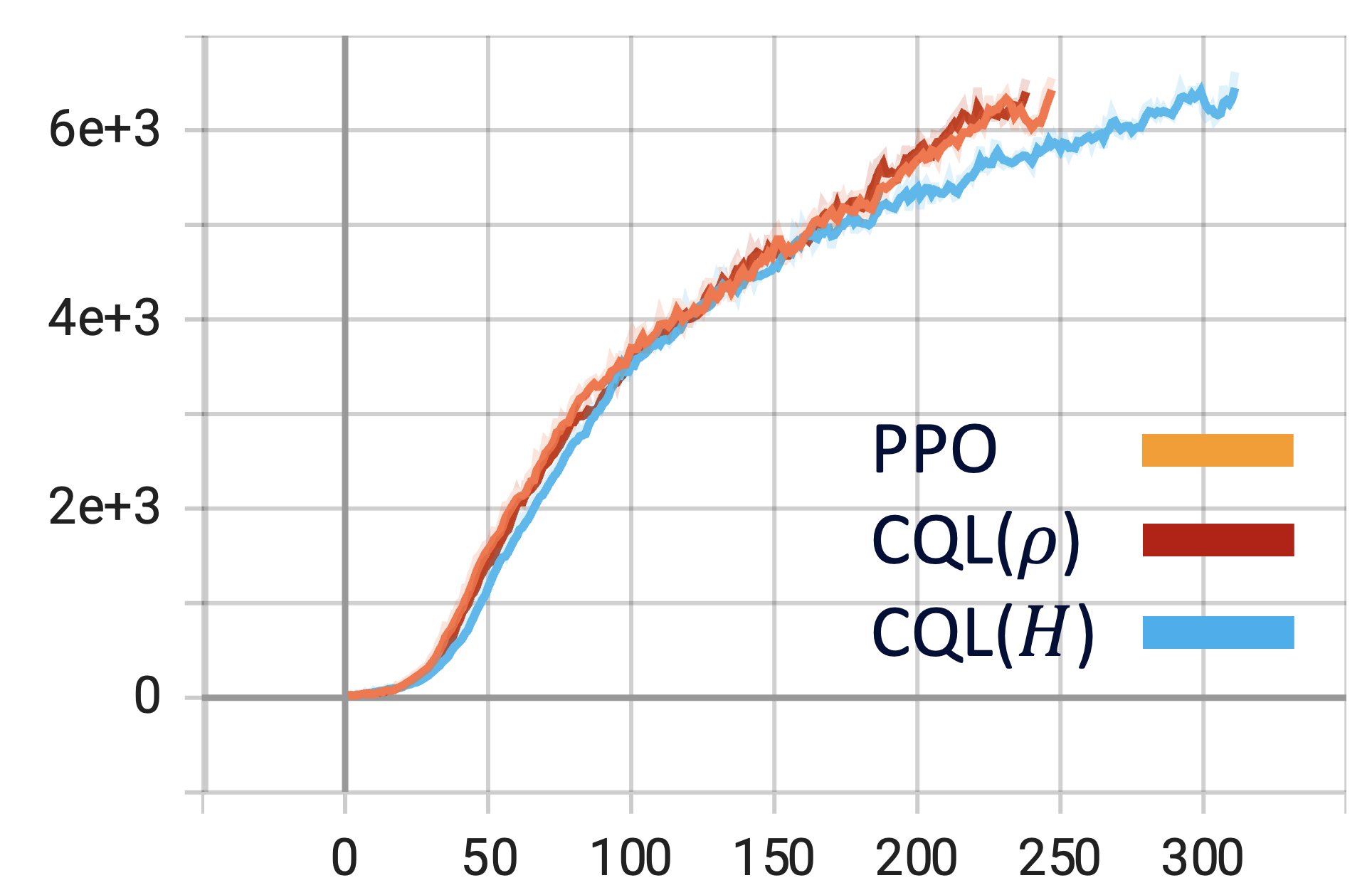}
             \caption{Ant}  
		    \label{ant_result}
         \end{subfigure}
         \hfill
         \begin{subfigure}[b]{0.3\textwidth}
             \centering
             \includegraphics[width=\textwidth]{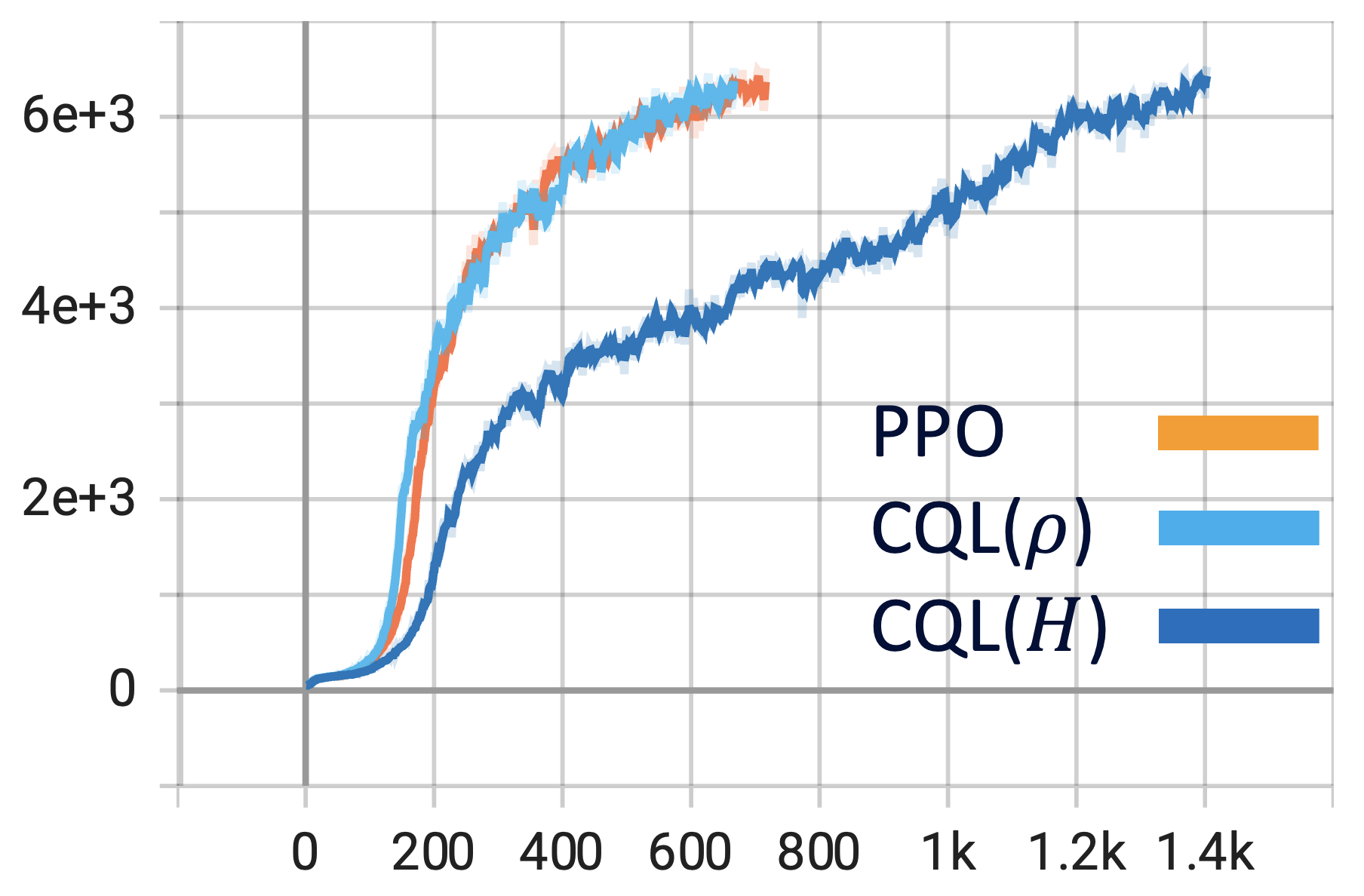}
             \caption{Humanoid}  
		\label{humanoid_result}
         \end{subfigure}
         \hfill
         \begin{subfigure}[b]{0.3\textwidth}
             \centering
             \includegraphics[width=\textwidth]{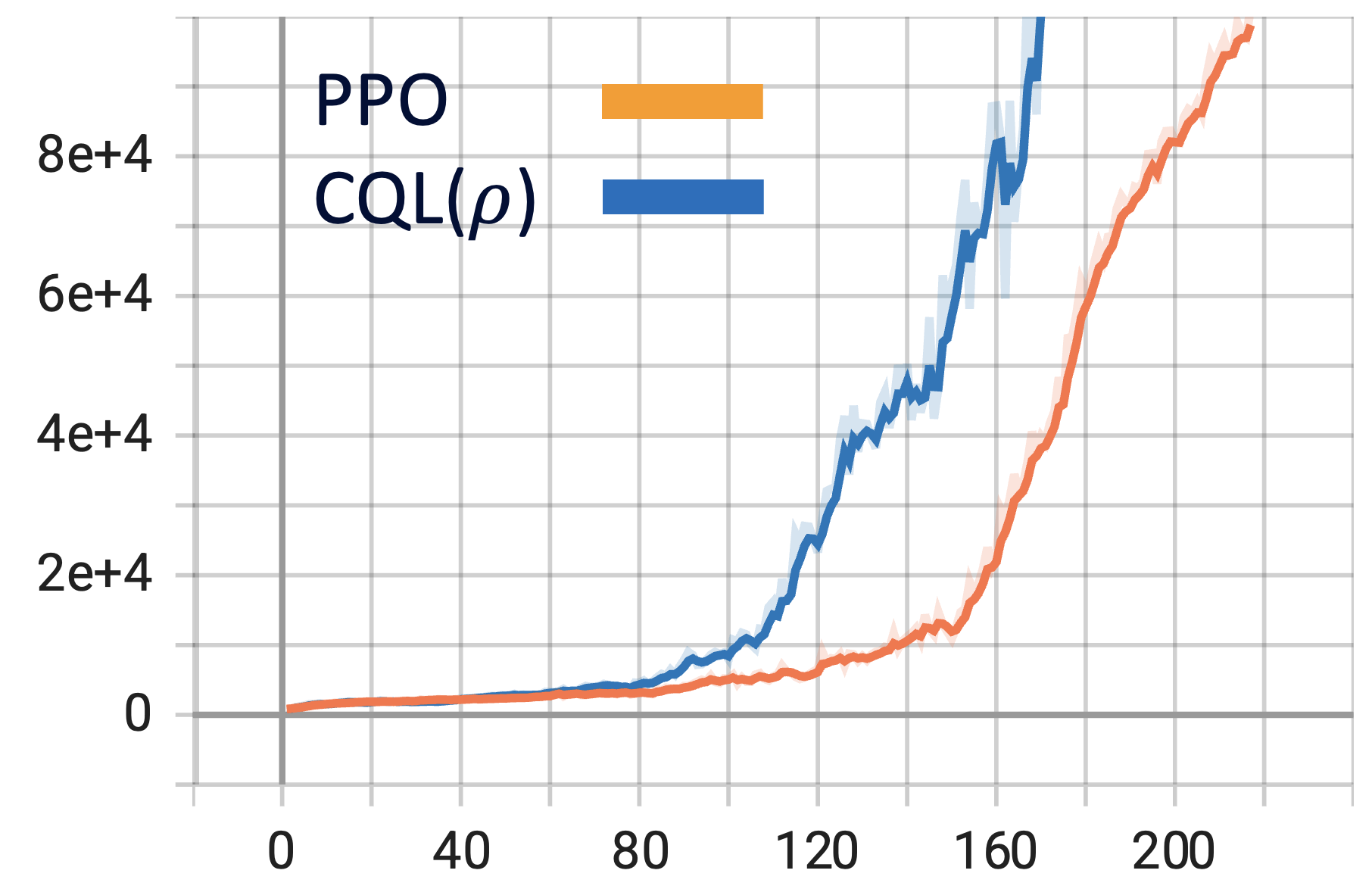}
             \caption{Our task}  
		\label{franka_result}
         \end{subfigure} \caption{Performance of PPO and CQL+PPO in different task environments} \label{threepics}
    \end{figure*}

    	We test the performance in Ant, Humanoid, and our task environments in general PPO method and variant CQL implement based on PPO, as shown in Fig. [\ref{ant_result}, \ref{humanoid_result}, \ref{franka_result}]. The results in Fig. \ref{threepics} show that CQL($\rho$) performs significantly better than the general PPO algorithm in our task. PPO and CQL act nearly the same in typical environments like Ant and Humanoid. This result is expected because, in the on-policy algorithm, the behavior policy is the same as the learned policy. Therefore, the effect of optimizing terms to promote learning efficiency is limited. Our \textit{Mixline} method effectively separates the whole long-horizon task into sub-tasks and can be easily extended to other skill-learning tasks.
	    
	\section{Conclusion}
	This paper proposes a novel hybrid reinforcement learning method for learning long-horizon and bimanual robot skills. To overcome the long-horizon issue, we combine the online and offline reinforcement learning algorithms as the \textit{Mixline} method to learn sub-task separately and then compose them together. Besides, we design a wait training mechanism to achieve bimanual coordination. The experiments are conducted in parallel based on the Isaac Gym simulator. The results show that using the \textit{Mixline} method can solve the long-horizon and bimanual coffee stirring task, which is intractable by just using online algorithms. The proposed method has the potential to be extended to other long-horizon and bimanual tasks. Moreover, combining online and offline RL algorithms might allow us to add human demonstration as the initial offline data to boost policy learning. Another further direction is to model the coordination mechanism by neural network rather than setting waiting manually.

\end{document}